\documentclass[letterpaper, 10 pt, conference]{ieeeconf}  

\IEEEoverridecommandlockouts                              

\usepackage{graphics,caption} 
\graphicspath{ {figure/} }
\usepackage{graphicx} 
\usepackage{epsfig} 
\usepackage{times} 
\usepackage{amsmath} 
\usepackage{amssymb} 
\usepackage{cite}
\usepackage{booktabs}
\usepackage{threeparttable}
\usepackage{multirow}
\if CLASSOPTIONcompsoc
\usepackage[caption=false, font=normalsize, labelfont=sf, textfont=sf]{subfig}
\else
\usepackage[caption=false, font=footnotesize]{subfig}
\usepackage{url}
\usepackage{mathrsfs}
\usepackage[colorlinks,linkcolor=blue]{hyperref}

\usepackage{float}
\usepackage{stfloats}

\title{\LARGE \bf
L3MVN: Leveraging Large Language Models for \\
 Visual Target Navigation\\
}

\author{Bangguo Yu, Hamidreza Kasaei, Ming Cao
\thanks{This work of Yu is supported in part by the China Scholarship Council.}
\thanks{All authors are with the Faculty of Science and Engineering, University of Groningen, 9747 AG Groningen, the Netherlands. {\tt\small \{b.yu, hamidreza.kasaei, m.cao\}@rug.nl}}%
}

\begin{document}

\maketitle
\thispagestyle{empty}
\pagestyle{empty}

\begin{abstract}

	Visual target navigation in unknown environments is a crucial problem in robotics.
	Despite extensive investigation of classical and learning-based approaches in the past, robots lack common-sense knowledge about household objects and layouts.
	Prior state-of-the-art approaches to this task rely on learning the priors during the training and typically require significant expensive resources and time for learning.
	To address this, we propose a new framework for visual target navigation that leverages Large Language Models (LLM) to impart common sense for object searching.
	Specifically, we introduce two paradigms: (i) zero-shot and (ii) feed-forward approaches that use language to find the relevant frontier from the semantic map as a long-term goal and explore the environment efficiently.
	Our analysis demonstrates the notable zero-shot generalization and transfer capabilities from the use of language.
	Experiments on Gibson and Habitat-Matterport 3D (HM3D) demonstrate that the proposed framework significantly outperforms existing map-based methods in terms of success rate and generalization. Ablation analysis also indicates that the common-sense knowledge from the language model leads to more efficient semantic exploration. Finally, we provide a real robot experiment to verify the applicability of our framework in real-world scenarios.
	The supplementary video and code can be accessed via the following link: \href{https://sites.google.com/view/l3mvn}{https://sites.google.com/view/l3mvn}.

\end{abstract}

\section{INTRODUCTION}

\begin{figure}[htbp]
	\centering
	\includegraphics[scale=0.6]{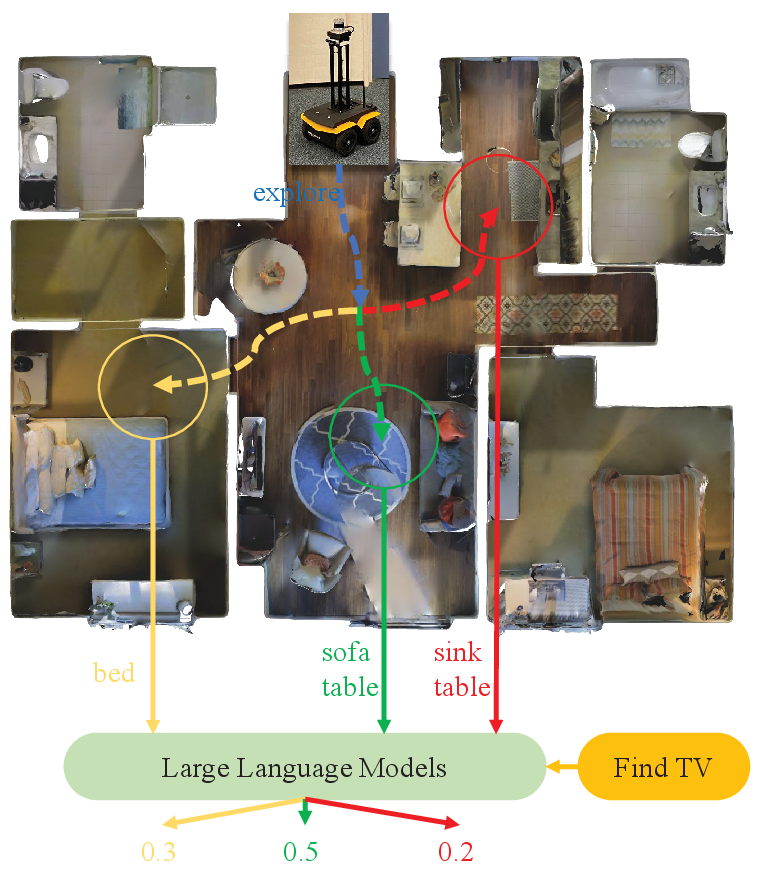}
	\caption{Visual target navigation example. The robot explores the environment and uses language models to find more relevant frontier (shown in green, with the highest score) based on the observation and the target.}
	\label{fig:demo}
	\vspace{-0.5cm}
\end{figure}

Human performance in complex and dynamic environments is partly attributed to their innate knowledge of the physical world, which is also crucial for intelligent agents in embodied tasks. 
Visual target navigation, a task that involves autonomous exploration in unknown environments, requires robots to possess semantic reasoning abilities.
In this task, the robot is provided with the name of an object, such as "toilet," and it must efficiently explore a 3D environment to locate any instance of the object.

The last few years have witnessed the development of large-scale photorealistic 3D scene datasets \cite{Ramakrishnan2021a}\cite{Xia2018}, fast simulators for embodied navigation\cite{Zhu2017}\cite{Savva2019}\cite{Xia2018}, reinforcement learning\cite{Mnih2013}\cite{Schulman2017}, and approaches for perception\cite{He2017}\cite{Jiang2018a}, cumulatively leading to huge progress for this task.
Traditional methods for robot navigation depend on the geometric maps building for path planning, which struggle to generalize to new instructions due to the lack of scene priors.
With the development of machine learning, learning-based methods directly optimize for navigation policies grounded in end-to-end or module methods, which gained close attention and made great progress on this task. Most of the learning-based methods formulate this task as a reinforcement learning (RL) problem and develop many useful presentations of the scene feature.
For the end-to-end models, the first framework\cite{Zhu2017} uses deep reinforcement learning to find an object as the appointed category in an unknown environment based on current observable images. Later, the framework is expanded by many researchers to improve the navigation performance\cite{Yang2019}\cite{Lyu2022}\cite{Druon2020}\cite{Ye2021a}.
Meanwhile, module-based methods \cite{Chaplot2020b}\cite{Chaplot2020}\cite{Ramakrishnan2022} attempt to use the explicit spatial map as scene memory and learns the semantic priors through deep reinforcement learning to determine the next step.

Visual target navigation methods that rely on learning typically demand substantial computational resources to acquire and employ scene priors. This motivates us to explore alternative approaches for obtaining scene priors that do not necessitate extensive learning processes.
Recent related works have explored interaction-free learning\cite{Ramakrishnan2022} and imitation learning\cite{Ramrakhya2022} to deal with this problem and significantly decrease the training cost. 
Another promising approach is to leverage large pre-trained models to transfer knowledge priors, such as zero-shot learning based on contrastive language image pretraining (CLIP) \cite{Radford2021}. While this approach shows potential, there is still significant room for improvement, as there remains a large gap between different tasks.



This study focuses on efficient navigation and searching for an object category in an environment based on visual observations.
Our proposed approach employs large language models to develop a policy for exploration and search, with language serving as a general tool for inferring relevance from observed objects. Specially, we leverage language to describe the contents of the frontiers in the map and employ language models to either perform zero-shot inference of which frontier best fits the description or embed the description as input for target-specific classifiers.
An illustrative example of visual target navigation for finding TV is shown in Fig \ref{fig:demo}. After scanning the environment, the robot need to select the next frontier to explore. Based on the objects around each frontier and our target object (tv), the large language model is adopted to find the most relevant frontier. 
Our model is evaluated on the Habitat simulation platform \cite{Savva2019} and compared to previous map-based navigation approaches. We conduct experiments on the photorealistic 3D environments of Gibson \cite{Xia2018} and HM3D \cite{Ramakrishnan2021a}. In sharp contrast to many other map-based methods, our framework uses the language models as knowledge priors for exploration direction selection, leading to reduced learning costs and improved the generalization of the scene priors. Ablation experiments demonstrate the effectiveness of the language model. Futhermore, we also showcase the application of our method in real-world settings.

Our contributions are summarized as follows:

\begin{itemize}

	\item We propose a framework that can build the environment map and select the long-term goal based on the frontiers with the inference of large language models to achieve efficient exploration and searching.
	\item We analysis two paradigms (i) zero-shot and (ii) feed-forward approaches for semantic reasoning about the objects using language models.
	\item We evaluate the performance of our model in the real-world setting using a robot platform and discuss the gap between simulation and reality for visual target navigation tasks.

\end{itemize}

\section{RELATED WORK}

\subsection{Visual Navigation}

Visual navigation is a key task for intelligent robots, inspired by human behavior. Classical approaches to visual navigation rely on building the environment map based on visual observation. These approaches plan a path using the map, which the robot follows to complete the navigation task. However, the accuracy of the environment map greatly affects navigation performance and classical methods struggle with unexplored or changing scenes, which require frontier-based exploration \cite{Yamauchi1997}. Recent works have proposed more visual navigation tasks, such as PointNav \cite{Wijmans2019}, ImageNav \cite{Zhu2017}, ObjectNav \cite{Chaplot2020b}, and visual-language navigation \cite{Huang2022a}. Here, we focus on the ObjectNav task in unknown environments using a novel framework.

\subsection{Navigational Policy Learning}



In the visual target navigation task, \cite{Zhu2017} proposed an end-to-end framework that used a pre-trained ResNet to encode the input observation and target image, which were then fused into an Asynchronous Advantage Actor-Critic (A3C) \cite{Mnih2013} model. To further enhance navigation performance, several methods have been proposed, such as knowledge graphs \cite{Yang2019}\cite{Lyu2022} or object relation graphs\cite{Druon2020}, large-scale training \cite{Wijmans2019}, human demonstrations \cite{Ramrakhya2022}, data augmentation \cite{Maksymets2021}, and auxiliary tasks \cite{Ye2021a}. These methods aim to improve the efficiency of the end-to-end navigational policy and enhance its generalization to novel scenes.

Compare to end-to-end methods that rely on past images \cite{Zhu2017}\cite{Yang2019}\cite{Druon2020} or RNN \cite{Lyu2022}\cite{Druon2020}\cite{Wijmans2019} to remember the scene features, map-based method has a strong ability to capture long-term dependencies using the map representation. These methods focus on the global policy of selecting waypoints on the environment map, and use classical path planners as the local policy to achieve navigation tasks. The global policy is learned from the different feature representations, such as topological graph\cite{Chaplot2020a}, geometry \cite{Chaplot2020} and semantic\cite{Chaplot2020b} map, or potential functions\cite{Ramakrishnan2022}. Recently, supervised learning has been used to learn potential functions and predict frontiers as long-term goals, resulting in improved performance and reduced computational costs for training, as demonstrated in \cite{Ramakrishnan2022}.

While deep reinforcement end-to-end and module-based techniques have achieved success in the visual navigation task, they still require learning scene priors from scene datasets. In contrast, our framework obtains scene priors from language models, enabling more general performance and reducing the learning process compared to prior work.

\subsection{Learning from Pretrained Models}

Recent advances in large pretrained vision and language models have shown their effectiveness in various domains, such as semantic segmentation\cite{Li2022}, scene understanding\cite{Chen2022}, and robot navigation\cite{Gadre2022}\cite{Huang2022a}.
Researchers have explored the use of these models for visual target navigation, including using CLIP\cite{Radford2021} as a visual enconder in RL\cite{Khandelwal2022}, object location\cite{Gadre2022}, and learning from other tasks \cite{Al-Halah2022}\cite{Min2022}. However, \cite{Gadre2022} only uses CLIP to locate the target object, and \cite{Khandelwal2022}\cite{Al-Halah2022}\cite{Min2022} still require learning navigational policies using reinforcement learning.
Our approach proposes a novel strategy for the semantic exploration module that leverages strong prior knowledge in language models, enabling zero-shot learning and fine-tuning-based learning for the visual target navigation task.

There are a number of recent works that relate to our approach, such as \cite{Ramakrishnan2022} and \cite{Chen2022}. \cite{Ramakrishnan2022} uses the potential functions with interaction-free learning to explores frontiers. \cite{Chen2022} achieves more general and zero-shot room classification using the priors of large language models. Our method specifically addresses the problem of learning a frontier-based navigation policy from large language models for visual target navigation. 
To the best of our knowledge, natural language processing tools have not been previously used for this task.

\section{The Proposed Method}

\begin{figure*}[htbp]
	\centering
	\includegraphics[scale=0.53]{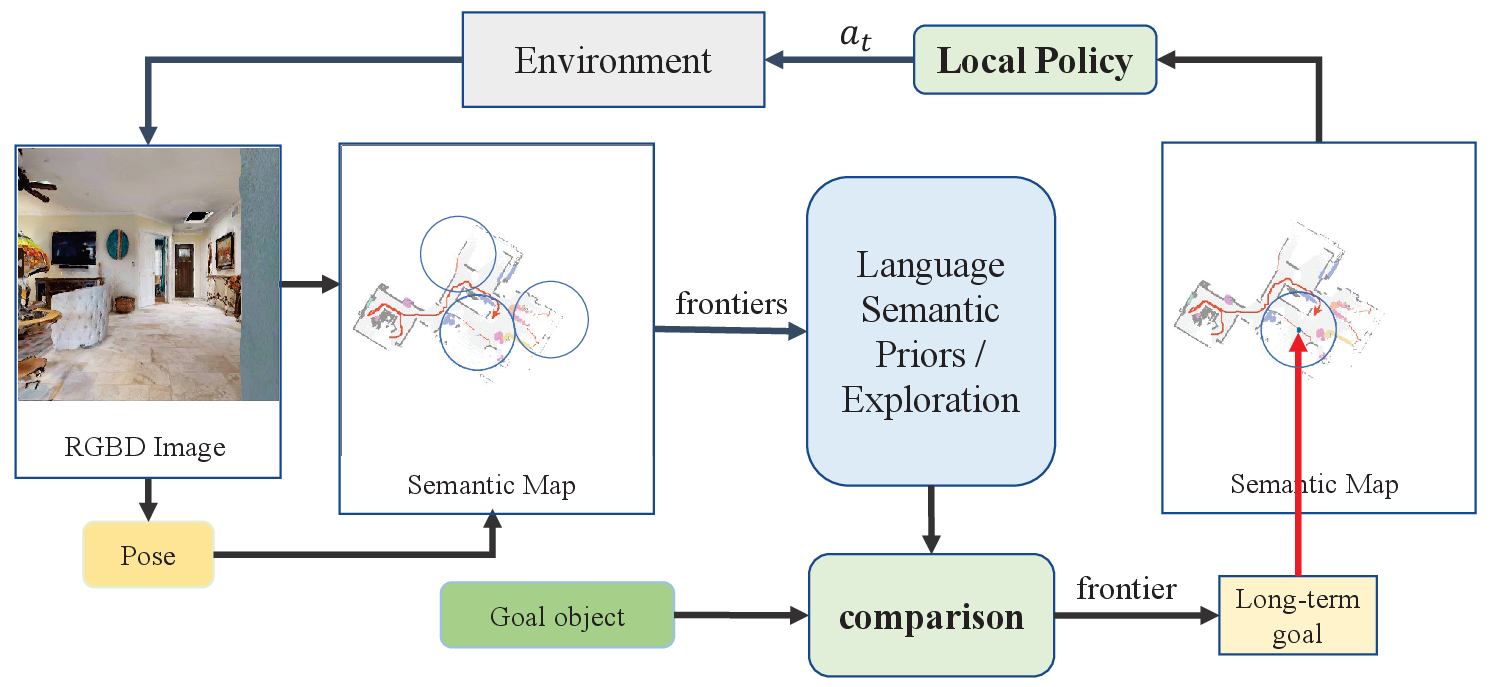}
	\caption{The architecture of the target navigation framework. The framework takes RGB-D images as input to generate a semantic map and frontiers, and selects a long-term goal based on the maps and object category using the inference of the language model. Once the long-term goal is reached, a local policy guides the final action for the robot.}
	\label{fig:system_architecture}
	\vspace{-0.4cm}
\end{figure*}

In this section, we describe the definition of the target navigation task and two paradigms with large language models.

\subsection{Task Definition}

The visual target navigation task involves the agent navigating an environment to find an object belonging to an appointed category. The category set is described by $C = \left\{c_0, \dots, c_m\right\}$, and the scene can be represented by $S = \left\{s_0, \dots, s_m\right\}$.
Each episode begins with the agent being initialized at a random position $p_i$ in the scene $s_i$, and receives the target object category $c_i$. Thus, an episode can be denoted as $T_i = \left\{s_i, c_i, p_i\right\}$.
At each time step $t$, the agent observes the environment and takes an action $a_t$. The observation includes RGB-D images, the agent's location and orientation, and the target object category. The action space, denoted by $\mathcal{A}$, includes six actions: $move\_forward$, $turn\_left$, $turn\_right$, $look\_up$, $look\_down$, and $stop$. The $move\_forward$ action agent moves 25 $cm$, while the $turn\_left$, $turn\_right$, $look\_up$, or $look\_down$ actions rotate the agent 30 degrees. The $stop$ action is used when the agent is close to the target object. If the agent takes the $stop$ action when the distance to the target is less than 0.1m, the episode is considered successful. The maximum number of time steps in an episode is 500.

\subsection{Overview}

Our framework is illustrated in Fig \ref{fig:system_architecture}.
Firstly, the agent obtains the observation of the environment to build the semantic map. The frontier map is extracted from the explored map and obstacle map.
Secondly, a long-term goal is selected from all the frontiers based on the relevance score between our target object and the observations. Large language models are used to infer the semantic relevance.
After getting the long-term goal, the local policy plans a path and takes the action to explore the environment and search for the target object.


\subsection{Map Representation}

\subsubsection{Semantic Map}




We construct the semantic map using RGB-D images and the agent's position, similar to \cite{Chaplot2020b}. It's represented as a $K \times M \times M$ tensor with $M \times M$ as map size and $K = C_n+2$ channels. The map is initialized with zeros at each episode's start, and the agent is placed at the center. Point clouds are generated from visual input using a geometric method, which is projected onto a top-down 2D map. The map contains obstacles and explored channels from depth images, and semantic channels from semantic segmentation. The semantic mask is aligned with the point clouds, and each channel is projected onto its corresponding position on the semantic map.

\subsubsection{Frontier Map}

We obtain the frontier map from the explored map and obstacle map, following the method in \cite{Ramakrishnan2022}. First, we extract the explored edge by identifying the maximum contours from the explored map. Then, by dilating the edge of the obstacle map, we generate the frontier map as the difference between the explored and obstacle maps. Next, we use connected neighborhoods to identify and cluster frontier cells in chains. We remove clusters that are too small. To score the frontier cells from the frontier map, we use the cost-utility approach proposed in \cite{Julia2012}. The subset of candidate destinations $F$ is composed of the cells in the center of the remaining cluster chains. So for each frontier cell $f \in F$, we can get the score $S^{C U}(a)$
\begin{equation}
	S^{C U}(a)=U(a)-\lambda_{C U} C(a)
\end{equation}
where $U(a)$ is a utility function, $C(a)$ is a cost function and the constant $\lambda_{C U}$ adjusts the relative importance between both factors.

\subsection{Global Policy}

After getting the semantic map and frontiers, we select a search window around each frontier and capture all the semantic objects as the frontier information, which are shown as blue circles in Figure \ref{fig:system_architecture}.
We present two paradigms to use the knowledge of the large language models and apply them for selecting of the frontiers. Both two paradigms summarize a frontier's contents in a query sentence, then process the sentence in the following ways:

\begin{itemize}
	\item \textbf{Zero-shot: } A pre-trained language model scores which object category is best described by the query.

	\item \textbf{Feed-forward: } The query string is embedded by a pre-trained language model. Then, a fine-tuned neural network outputs a distribution over object categories given that embedding.
\end{itemize}

\begin{figure}[htbp]
	\centering
	\includegraphics[scale=0.35]{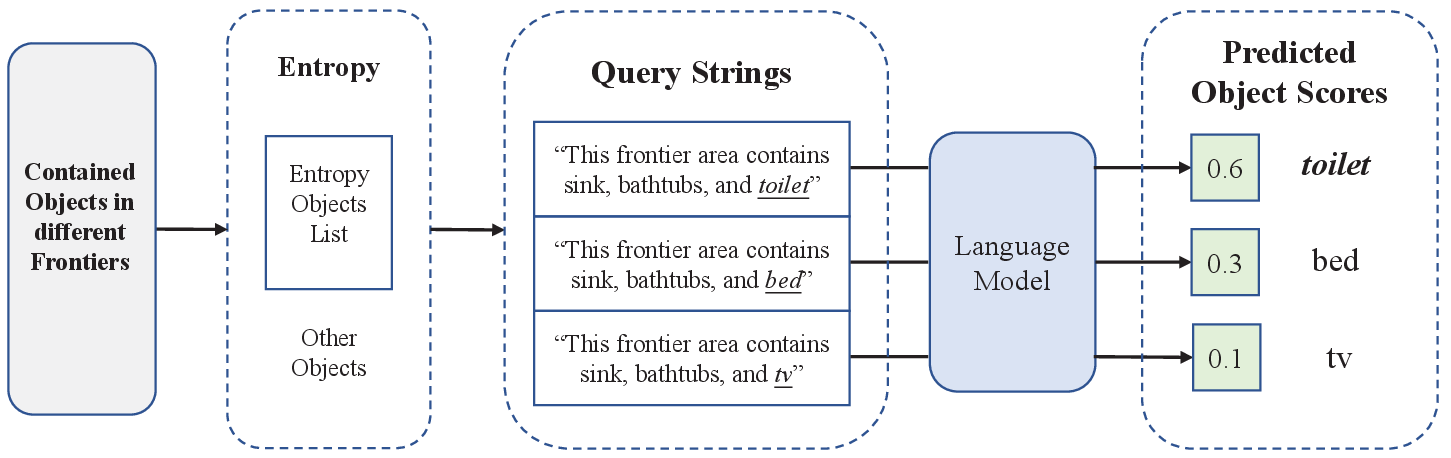}
	\caption{Example of zero-shot approach.}
	\label{fig:zeroshot}
\end{figure}
\begin{figure}[htbp]
	\centering
	\includegraphics[scale=0.35]{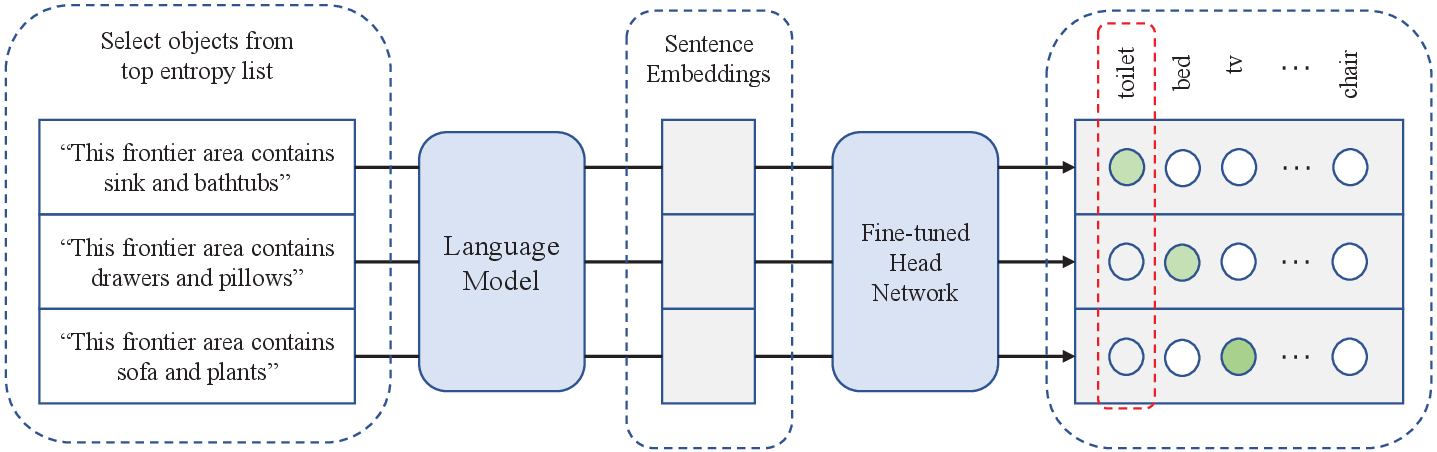}
	\caption{Example of the fine-tuning-based feed-forward approach.}
	\label{fig:ff}
\end{figure}

\subsubsection{Preprocessing for Language Model}

We want to use a sentence to describe the semantic information around the frontier, and then use the language model to score the description. 
To assess the relevance between a frontier and a target object, \textit{masked language models} (MLMs) is used to score a string $W$ based on semantic and grammatical sensibility as \cite{Chen2022}, such as the score of the sentence ``This frontier area contains sink, bathtubs, and toilet.'' is higher than the sentence ``This frontier area contains sink, bathtubs, and tv.'' Based on the scores of strings containing common-sense facts, a proxy measure can be obtained for how likely it is for the fact to be true. Additionally, we utilize the high-dimensional embeddings of the language model, denoted as $Emb(W)$, to represent the meaning and grammatical structure of the text. By fine-tuning the language model, we aim to learn a mapping from the embeddings to a prediction score for each target object, thus providing a quantitative measure of the relevance between the frontier and the object.


To account for the complexity of the environment, we calculate the entropy of each object category, as proposed in \cite{Chen2022}, in order to mitigate the influence of uninformative and ubiquitous objects (e.g., doors and windows). Specifically, objects that appear less frequently across the target object categories are deemed more informative, as their presence tends to imply certain target categories. This is reflected in their non-uniform conditional distributions $p(t_j | o_i)$, where $o_i \in L_O$ represents the object category, $t_j \in L_T$ represents the target category, and $L_T$ and $L_O$ are the sets of target object and object categories, respectively. In our task, $L_T \subseteq L_O$ implies that some target objects may also be informative object categories. We compute these conditional probabilities using ground-truth co-occurrences from publicly available semantic annotated scene datasets \cite{Ramakrishnan2021a} containing of 1,000 high-resolution 3D scans of indoor spaces. Specifically, we count the number of times each object category appears around each target category, and normalize the counts over targets to obtain $p(t_j | o_i)$. Once we have access to $p(t_j | o_i)$, we can calculate the entropy using the following formula:

\begin{equation}
	H_{O_i} = -\sum_{t_j \in L_T} p(t_j | o_i) \log p(t_j | o_i)
\end{equation}
Entropy is maximized when the considered distribution is uniform and minimized when it is one-hot, meaning more semantically-informative objects have lower corresponding entropy. We create a top entropy list of the 15 lowest-scoring objects and restrict the composition of the input string $W$ to only include objects from this list, which differs from \cite{Chen2022}. We use the resulting string to infer relevance between the frontier and target objects.

\subsubsection{Zero-shot Approach}
For the zero-shot approach, we construct $|L_T|$ query strings for each frontier $f_i$, one per target category:
\begin{equation}
	\begin{aligned}
		 & W_{t_j}^{f_i} = \text{``A frontier area containing }            \\
		 & o_1, o_2, \cdots, o_k, t_j \text{.'' } \forall t_j \in L_T
	\end{aligned}
\end{equation}
where $o_{1\cdots k}$ are the objects detected from the frontier area $f_i$, and $t_j$ is the target object. All these queries are scored via language model, with the final estimated frontier $f$ being whichever one yields the highest query sentence probability for the target $t_j$. The score for each frontier is:

\begin{equation}
	S^{LLM}_{f_i} = \log p(W_{t_j}^{f_i})
\end{equation}

\subsubsection{Feed-forward Approach}

For the feed-forward approach, we create a single query string of the form for each frontier $f_i$:
\begin{equation}
	W^{f_i} = \text{``This frontier contains } o_1, \cdots, \text{ and } o_k \text{.''}
\end{equation}
This string is then fed into a language model to produce a summary embedding vector. Finally, the embedding is fed into a fine-tuned neural network head, which produces a $|L_T|$-dimensional vector of prediction logits corresponding to the target categories, with the inferred frontier being the one corresponding to the maximum value of the target object's output. The score for each frontier is:
\begin{equation}
	S^{LLM}_{f_i} = [f_\theta (Emb(W^{f_i}))]_{t_j}
\end{equation}
where $f_\theta: Emb(W^{f_i}) \rightarrow \mathbb{R}^{L_T}$ is a neural network that takes in query embeddings and outputs prediction logits. We use three-layer network for this mapping.

\subsubsection{Exploration}

The sparsity of semantic information in indoor environments can result in frontier areas that lack informative objects or contain unrelated objects (e.g., a sofa in a toilet). To address this, we integrate the frontier map score $S^{CU}$ with the language model score $S^{LLM}$. We set a score bound $B$ for the frontier and normalize $S^{CU}$ to fit within this range. We then select the frontier using the following strategy:
\begin{equation}
	f =\arg \max_{f_i} \left\{\begin{array}{cc}
		S^{LLM}_{f_i},           & \text{if } \max[S^{LLM}] > \sup(B)  \\[3mm]
		(S^{LLM}_{f_i}, S^{CU}_{f_i}), & \text{if } \max[S^{LLM}] \in B  \\[3mm]
		S^{CU}_{f_i},            & \text{if }  \max[S^{LLM}] < \inf(B)
	\end{array}\right.
\end{equation}

\subsection{Local Policy}

In order to navigate from the agent's current location to a long-term goal, we employ the Fast Marching Method (FMM) \cite{Sethian1996}. Subsequently, the agent selects a local goal within a restricted range of its current position and executes the final action $a_t \in \mathcal{A}$ to reach it. At each step, the local map and local goal are updated based on new observations. This approach, which employs module policies, enhances training efficiency and obviates the necessity of learning obstacle avoidance from scratch, as is required in the end-to-end approach.

\renewcommand\arraystretch{1.4}
\begin{table*}[htbp]
	\centering
	\fontsize{9}{8}\selectfont
	\begin{threeparttable}
		\caption{Results of Comparative Study.}
		\label{tab:performance_comparison}
		\setlength{\tabcolsep}{4mm}{}
		{
			\begin{tabular}{ccccccc}
				\toprule
				\multirow{2}{*}{Method}                   & \multicolumn{3}{c}{ Gibson } & \multicolumn{3}{c}{ HM3D }  \cr
				\cmidrule(lr){2-4} \cmidrule(lr){5-7}
				                                          & Success                      & SPL                             & DTG         & Success     & SPL         & DTG\cr
				\midrule
				Random Walking                            & 0.030                        & 0.030                           & 2.580       & 0.000       & 0.000       & 7.600          \cr
				Frontier Based Method \cite{Yamauchi1997} & 0.417                        & 0.214                           & 2.634       & 0.237       & 0.123       & 5.414             \cr
				Random Sample on Map                      & 0.544                        & 0.288                           & 1.918       & 0.300       & 0.143       & 4.761          \cr
				SemExp \cite{Chaplot2020b}                & 0.652                        & 0.336                           & 1.520       & 0.379       & 0.188       & 2.943             \cr
				PONI \cite{Ramakrishnan2022}              & 0.736                        & 0.410                           & 1.250       & -           & -           & -       \cr
				L3MVN (Zero-Shot)                         & {\bf 0.761}                  & {\bf 0.377}                     & {\bf 1.101} & {\bf 0.504} & {\bf 0.231} & {\bf 4.427}       \cr
				L3MVN (Feed-forward)                      & {\bf 0.769}                  & {\bf 0.388}                     & {\bf 1.008} & {\bf 0.542} & {\bf 0.255} & {\bf 3.934}       \cr
				\bottomrule
			\end{tabular}
		}

	\end{threeparttable}
\end{table*}

\begin{figure*}[htbp]
	\centering
	\subfloat[step 10]
	{
		\includegraphics[width=4.2cm]{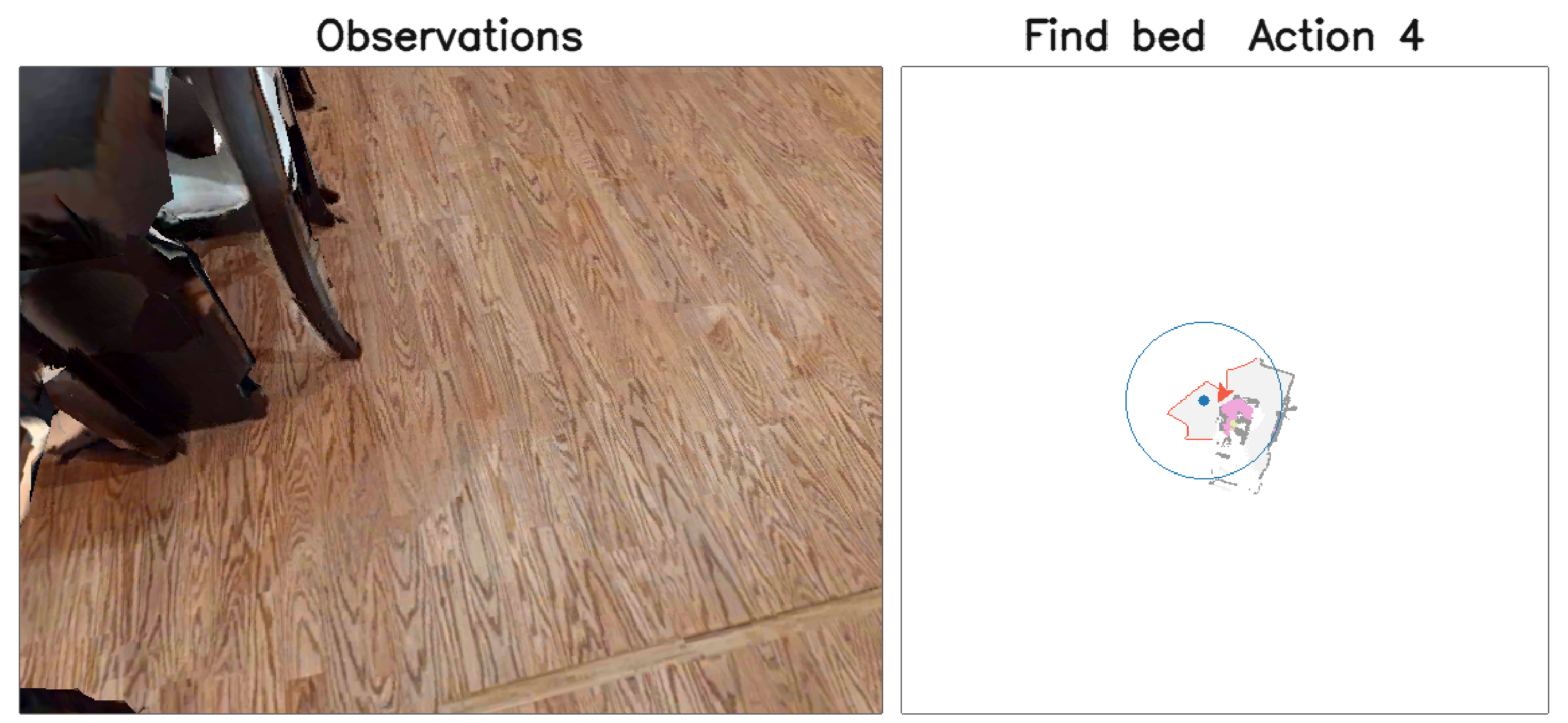}
	}
	\subfloat[step 60]
	{
		\includegraphics[width=4.2cm]{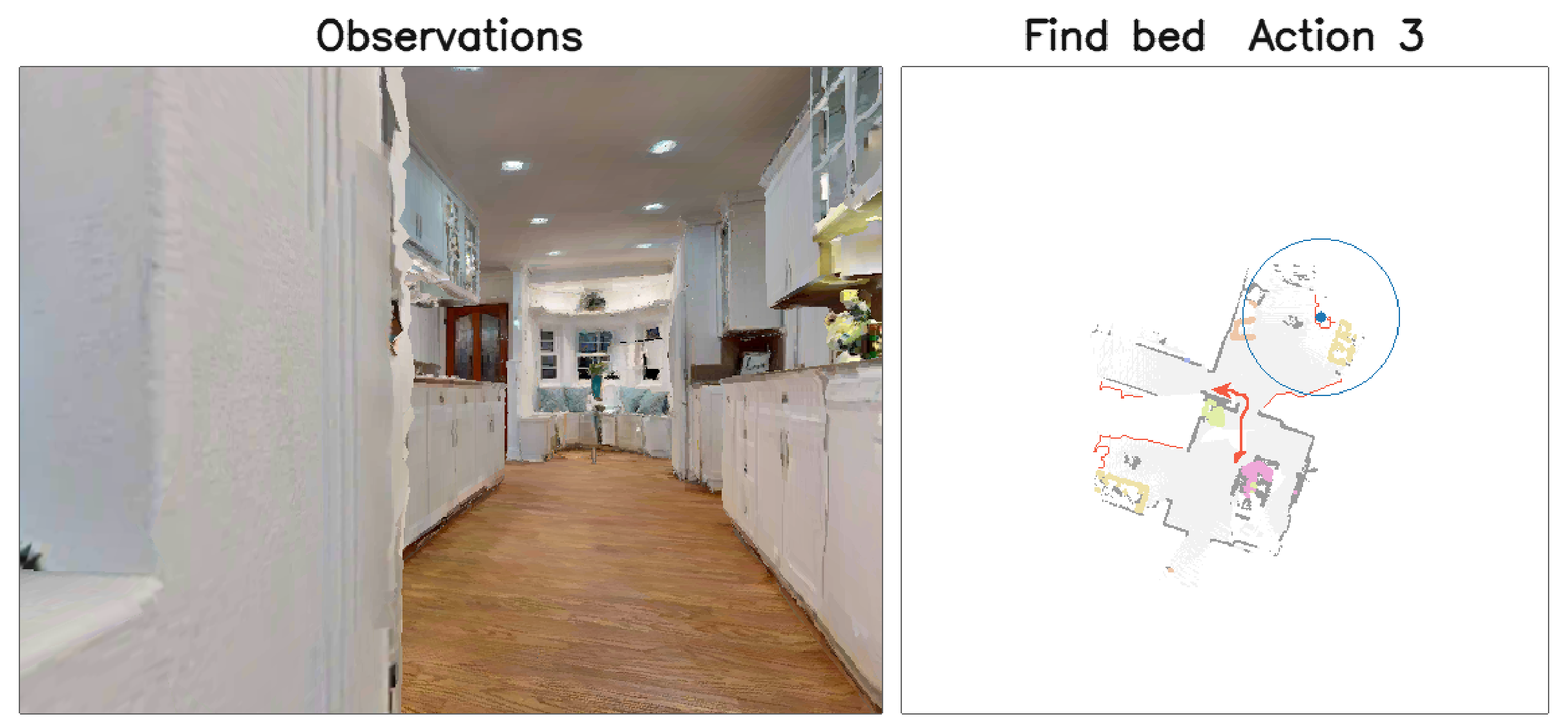}
	}
	\subfloat[step 110]
	{
		\includegraphics[width=4.2cm]{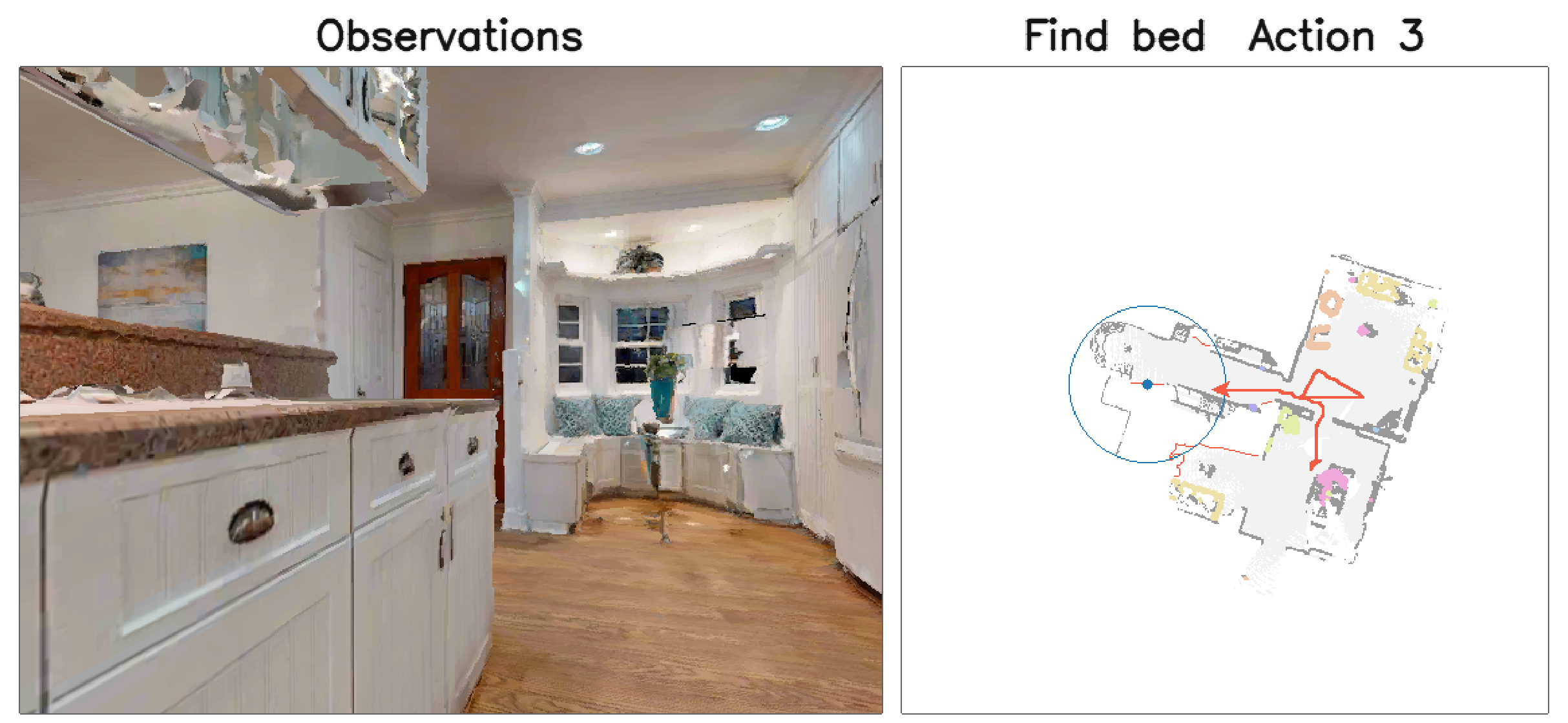}
	}
	\subfloat[step 169]
	{
		\includegraphics[width=4.2cm]{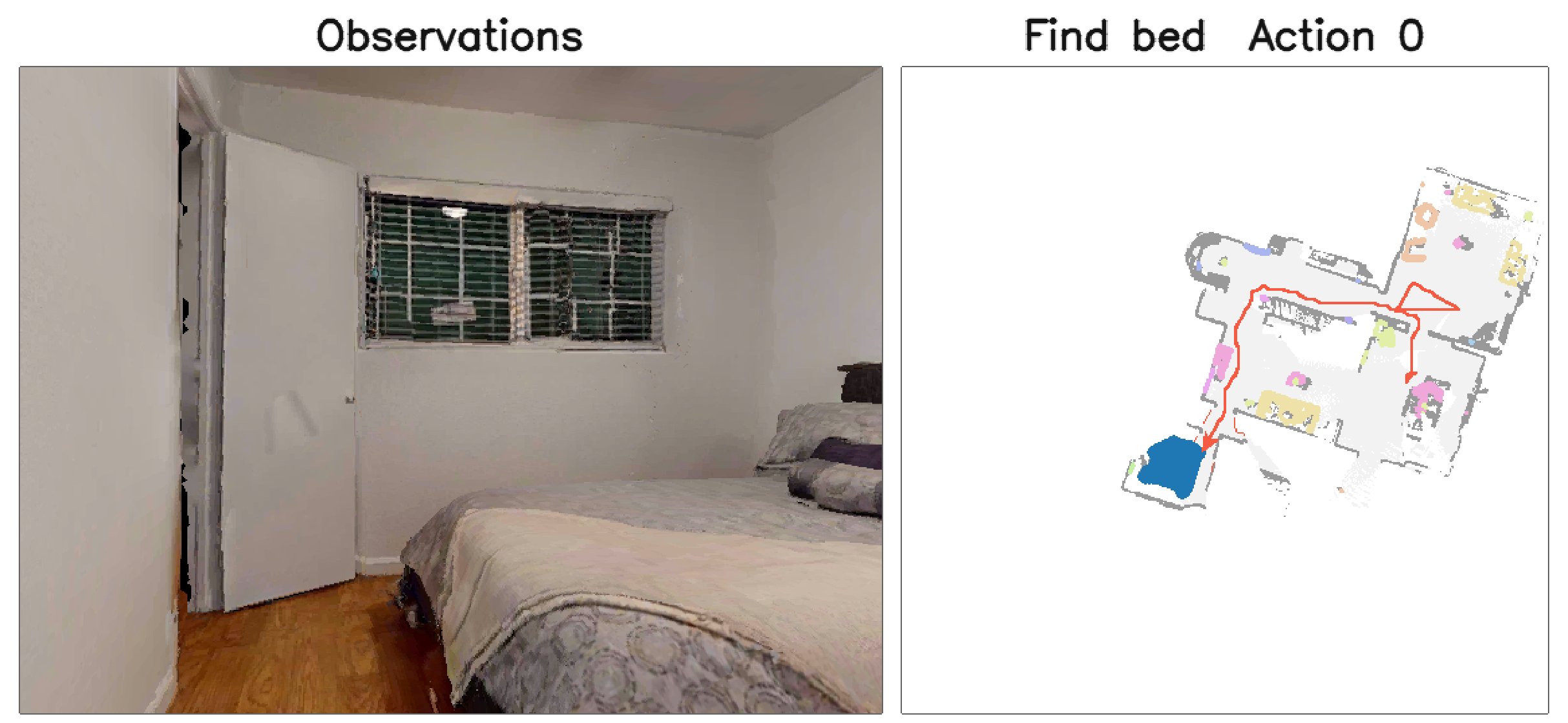}
	}
	\caption{The visual target navigation experiment process in the Habitat platform for finding a bed. The gray channel represents the barrier, the blue spot and the circle denote the long-term goal selected by our policy, the red thick line represents the trajectory of the robot, the red thin line denotes the frontiers, and other colors represent the semantic objects.}
	\label{fig:sim_episode}
	\vspace{-0.4cm}
\end{figure*}

\section{EXPERIMENTS}

\begin{figure*}[htbp]
	\centering
	\subfloat[step 10]
	{
		\includegraphics[width=4.2cm]{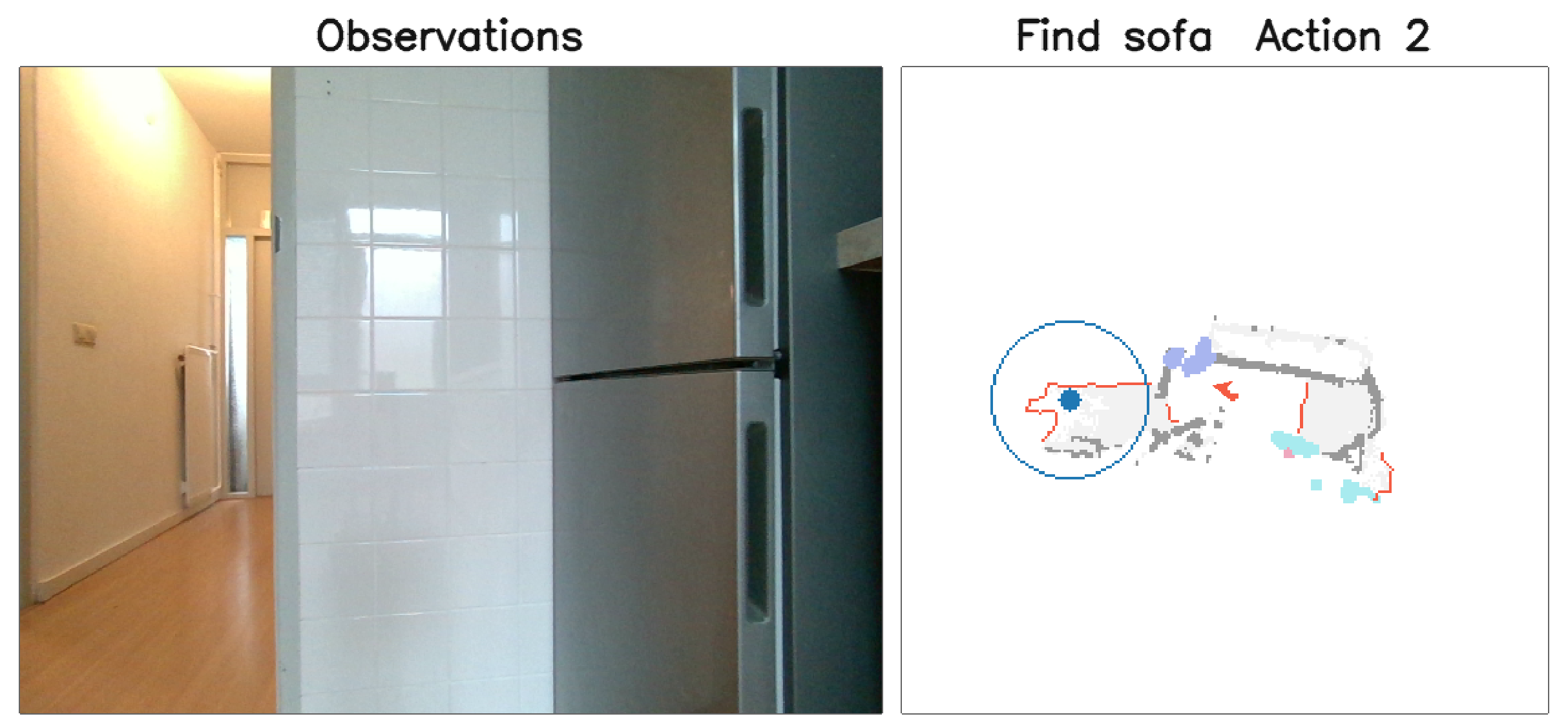}
	}
	\subfloat[step 25]
	{
		\includegraphics[width=4.2cm]{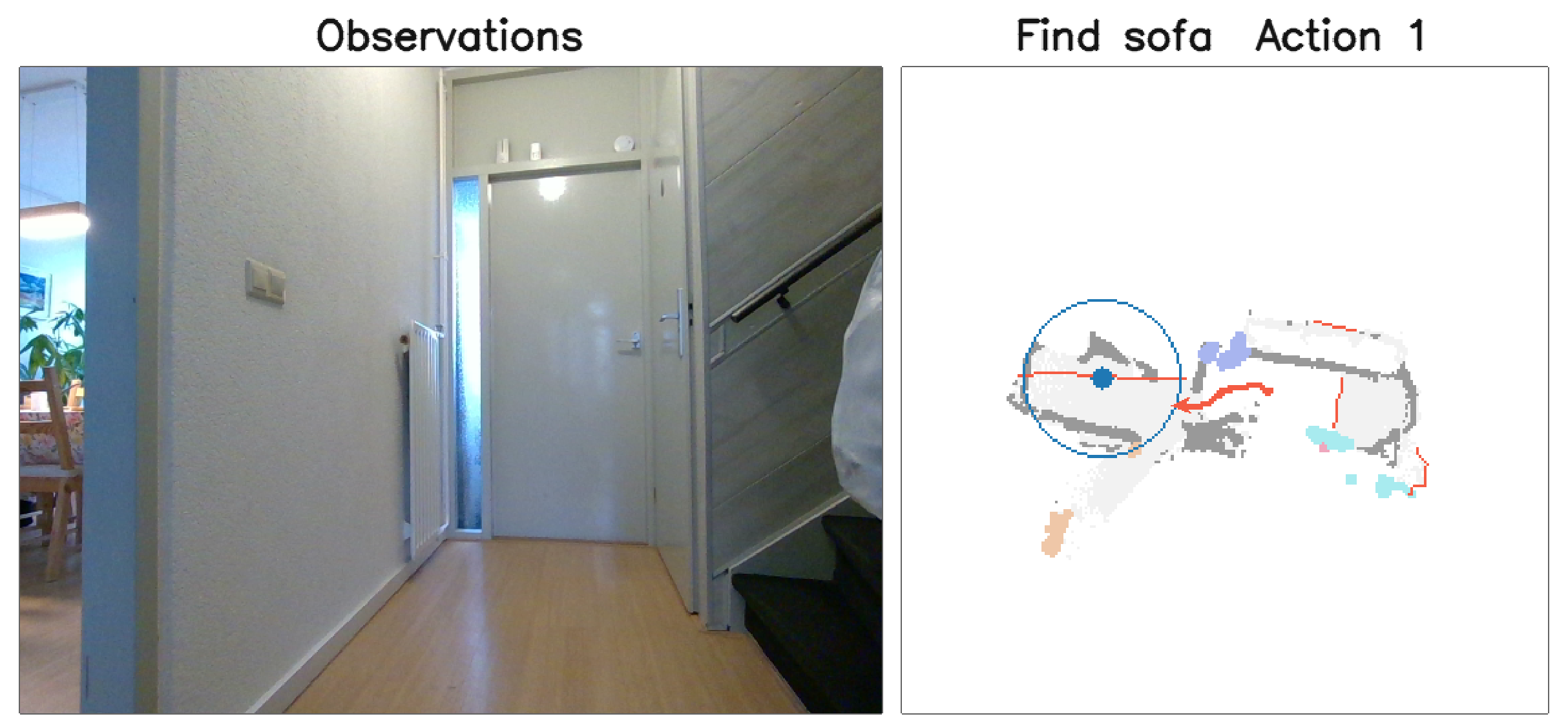}
	}
	\subfloat[step 40]
	{
		\includegraphics[width=4.2cm]{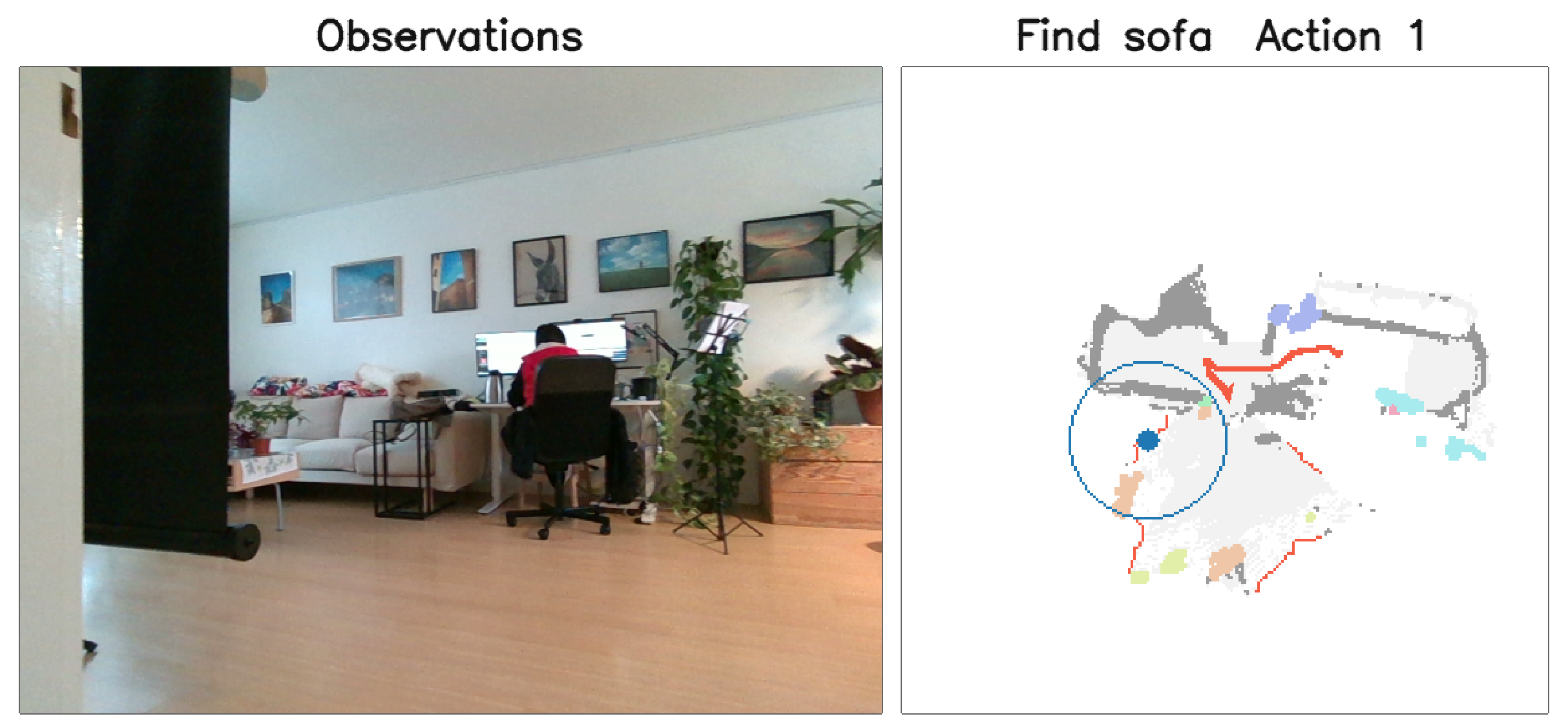}
	}
	\subfloat[step 58]
	{
		\includegraphics[width=4.2cm]{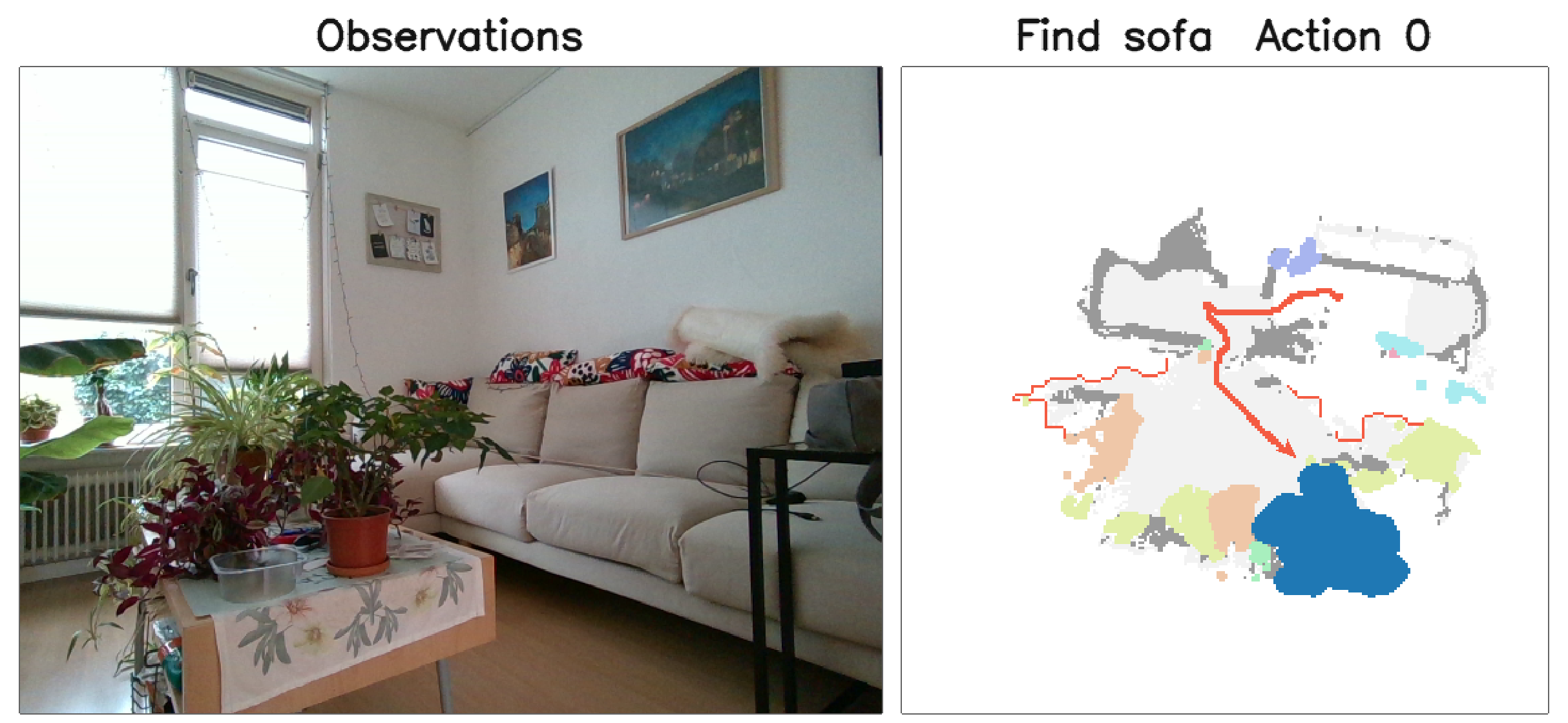}
	}
	\caption{The process of visual target navigation experiment in the real world to find a sofa.}
	\label{fig:real_episode}
	\vspace{-0.4cm}
\end{figure*}

In this section, we evaluate the performance of our method by comparing it with other map-based baselines in a simulated environment. Additionally, we apply our method in a real-world robot platform to validate its practicality for navigational tasks.

\subsection{Simulation Experiment}
\subsubsection{Dataset}

Our experiments are conducted on two high-resolution photorealistic 3D reconstructions of real-world environments: HM3D \cite{Ramakrishnan2021a} and Gibson \cite{Xia2018}.
The Gibson dataset comprises 25 training and 5 validation scenes from the Gibson tiny split, which come with associated semantic annotations. The HM3D dataset is in the habitat format, and we use the standard splits of 75 training and 20 validation scenes.
There are 6 object goal categories defined \cite{Chaplot2020b}: chair, couch, potted plant, bed, toilet, and TV.


\subsubsection{Experiment Details}
\label{sec:zerovsfeed}
We conducted our evaluation on the 3D indoor simulator Habitat platform \cite{Savva2019}, using an observation space consisting of $480\times640$ RGBD images, a base odometry sensor, and a goal object represented as an integer. We utilized the finetuned RedNet model \cite{Jiang2018a} to predict all categories as in \cite{Ye2021a}. Our implementation was based on publicly available code from \cite{Chaplot2020b} and \cite{Chen2022}, using the PyTorch framework. We set the score range for the frontier to be between 0 and 1, with a bound of $B = [0.15, 0.3]$.

\begin{itemize}
	\item For the zero-shot approach, we utilized the pre-trained language model RoBERTa-large \cite{Liu2019} to evaluate the query strings that were generated based on the semantic observation around each frontier and the target object.
	\item For the feed-forward approach, we finetune RoBERTa-large as our sentence embedder and train head networks on HM3D datasets. We extract all the objects in each room of the HM3D scenes and check if the room contains the target objects. If so, we split the target object as the label and other objects as the sample to create the dataset. Finally, we train the finetuned models with RoBERTa embeddings on the entire dataset and evaluate the query strings for each frontier.
\end{itemize}


\subsubsection{Evaluation Metrics}

We follow \cite{Anderson2018} to evaluate our method using Success Rate, Success weighted by Path Length (SPL), and Distance to Goal (DTG). SR is defined as $\frac{1}{N} \sum_{i=1}^{N} S_{i}$, and SPL is defined as $\frac{1}{N} \sum_{i=1}^{N} S_{i} \frac{l_{i}}{\max \left(l_{i}, p_{i}\right)}$, where $N$ is the number of episodes, $S_{i}$ is 1 if the episode is successful, else is 0, $l_{i}$ is the shortest trajectory length between the start position and one of the success position, $p_{i}$ is the trajectory length of the current episode $i$. The DTG is the distance between the agent and the target goal when the episode ends.


\subsubsection{Baselines}

In order to evaluate the navigation performance of our model, we considered several baselines.

\begin{itemize}

	\item \textbf{Randomly Walking: }At each step, the agent selects an action uniformly at random from the action space $\mathcal{A}$.
	\item \textbf{Frontier-based Policy \cite{Yamauchi1997}: }This baseline method employs a classical robotics pipeline for mapping and a frontier-based exploration algorithm. 
	\item \textbf{Randomly Sampling on Map: }We randomly sample the long-term goal from the map, and use the local planner method to select the final action.
	\item \textbf{SemExp \cite{Chaplot2020b}: }We follow \cite{Chaplot2020b} as the baseline to explore and search for the target using semantic map.
	\item \textbf{PONI \cite{Ramakrishnan2022}: }Potential function \cite{Chaplot2020b} is the newest map-based work and can be set as baseline of interaction-free learning method. We can only get the results on Gibson datasets from the published work.

\end{itemize}


\subsubsection{Result and Discussion}

The quantitative results of the comparison study are reported in TABLE \ref{tab:performance_comparison}.
As indicated by the results, random walking without any specialized navigation policy fails in almost all episodes.
However, when we utilize the map-based framework to randomly sample the long-term goal, the performance is even superior to that of the classical frontier-based method \cite{Yamauchi1997}. This indicates the significant advantage of the map-based method in allowing the robot to quickly and roughly explore the environment.
Furthermore, the significant improvement achieved by SemExp \cite{Chaplot2020b} highlights the importance of semantic information in efficient exploration.
PONI \cite{Ramakrishnan2022} further enhances the performance and reduces computational costs, demonstrating the capacity for learning semantic priors in a distinct way from other reinforcement learning approaches.
Our framework consistently outperforms all baselines across both datasets, achieving notable improvement over the SemExp \cite{Chaplot2020b} and PONI \cite{Ramakrishnan2022} baselines. 
The comparison between the feed-forward and zero-shot approaches indicates that feed-forward method learns more accurate relevance in large indoor scenes.
The process of finding a bed is illustrated in Fig \ref{fig:sim_episode}.

\renewcommand\arraystretch{1.4}
\begin{table}[htbp]
	\centering
	\fontsize{8}{8}\selectfont
	\begin{threeparttable}
		\caption{Results of Ablation Study in HM3D.}
		\label{tab:ablation_study}
		\setlength{\tabcolsep}{2mm}{}
		{
			\begin{tabular}{ccccccc}
				\toprule
				\multicolumn{4}{c}{L3MVN ablation} & \multicolumn{3}{c}{ HM3D results }\cr
				\cmidrule(lr){1-4} \cmidrule(lr){5-7} 
				Near                               & Exp                                & LLM                                    & GT Seg       & Success$\uparrow$  & SPL$\uparrow $   & DTG$\downarrow$    \cr
				\midrule
				$\checkmark$                       &                                    & $\checkmark$                           &              & 0.490   & 0.231 & 4.335           \cr
				                                   & $\checkmark$                       &                                        &              & 0.518   & 0.244 & 4.084             \cr
				                                   & $\checkmark$                       & $\checkmark$                           &              & 0.542   & 0.255 & 3.934          \cr
				                      & $\checkmark$                       & $\checkmark$                           & $\checkmark$ & 0.664   & 0.364 & 3.195             \cr
				\bottomrule
			\end{tabular}
		}

	\end{threeparttable}
	\vspace{-0.4cm}
\end{table}

\subsubsection{Ablation study}

To assess the relative importance of the various modules within our framework, we have performed the ablations using the HM3D dataset: the cost-utility exploration module (Exp), the feed-forward approach for language module (LLM), the nearest frontier module (Near) that only used for LLM ablation, and the ground-truth semantic segmentation (GT Seg). The results in TABLE \ref{tab:ablation_study} show that our complete model achieves the best performance (row 3, TABLE \ref{tab:ablation_study}), and the LLM and exploration policy are crucial for achieving good performance. Removing the LLM (row 2, TABLE \ref{tab:ablation_study}) decreases both success rate and SPL but still higher than the zero-shot approach, and replacing cost-utility exploration with the nearest frontier policy (row 1, TABLE \ref{tab:ablation_study}) leads to a further drop in performance, which shows the importance of exploration for this task. Augmenting our complete model with GT Seg improves performance on all cases, as semantic segmentation impacts the semantic mapping, which is also the main reason for the failures.
\subsection{Real World Experiment}

We utilized ROS and the Jackal Robot hardware platform with Realsense D455 camera and Ouster lidar to implement and test our policy in the real world. To minimize the gap between the simulation and the real-world scenarios, we attempted to keep the sensor data consistent with the simulation environment. Specifically, we set the same height and range of the depth image for the robot's RGB-D camera as in the simulation environment, and used the lidar to construct a real-time geometric map to improve the robot's localization accuracy. The RGB-D images, location, and object category were then fed into our model, which produced an action output. As shown in Fig \ref{fig:real_episode}, the long-term goal selected by our model (represented by a blue spot) guided the robot to efficiently explore the environment and locate the goal.

The RGB-D images and location are the primary differences between the simulation and the real world, as the depth image is affected by factors such as illumination and hardware quality. Furthermore, the lidar and odometry measurements in the real world are not as accurate as in the simulation, leading to noisy points around objects and walls. 
But the module-based framework has the advantage in real-world transfer since it takes the scene map as input rather than direct images with noise. This reduces the impact of noise, allowing us to deploy the model on the robot platform with minimal fine-tuning.

\section{CONCLUSIONS}

We presented L3MVN, a novel module framework that leverages large language models to facilitate visual target navigation by examining two paradigms that infer the semantic relevance from the observed frontiers. By implementing experiments in Gibson and HM3D datasets, we demonstrate that our approach significantly improves the success rate and efficiency while avoiding large-scale learning processes. Ablation studies demonstrated that our language models and cost-utility exploration lead to more efficient navigation. We also validate the effectiveness in a real-world experiment, highlighting the practical applicability of our method. Our findings suggest that large language models hold immense potential in aiding robots in such tasks by providing useful knowledge. Future research should consider the design of the interaction between the robot and LLM.






\bibliographystyle{ieeetr}
\bibliography{bib/library.bib}

\begin{thebibliography}{10}

\bibitem{Ramakrishnan2021a}
S.~K. Ramakrishnan, A.~Gokaslan, E.~Wijmans, O.~Maksymets, A.~Clegg, J.~Turner,
  E.~Undersander, W.~Galuba, A.~Westbury, A.~X. Chang, M.~Savva, Y.~Zhao, and
  D.~Batra, ``{Habitat-Matterport 3D Dataset (HM3D): 1000 Large-scale 3D
  Environments for Embodied AI},'' {\em arXiv}, sep 2021.

\bibitem{Xia2018}
F.~Xia, A.~R. Zamir, Z.~He, A.~Sax, J.~Malik, and S.~Savarese, ``{Gibson Env:
  Real-World Perception for Embodied Agents},'' in {\em 2018 IEEE/CVF
  Conference on Computer Vision and Pattern Recognition}, pp.~9068--9079, IEEE,
  jun 2018.

\bibitem{Zhu2017}
Y.~Zhu, R.~Mottaghi, E.~Kolve, J.~J. Lim, A.~Gupta, L.~Fei-Fei, and A.~Farhadi,
  ``{Target-driven visual navigation in indoor scenes using deep reinforcement
  learning},'' in {\em Proceedings - IEEE International Conference on Robotics
  and Automation}, pp.~3357--3364, IEEE, may 2017.

\bibitem{Savva2019}
M.~Savva, A.~Kadian, O.~Maksymets, Y.~Zhao, E.~Wijmans, B.~Jain, J.~Straub,
  J.~Liu, V.~Koltun, J.~Malik, D.~Parikh, and D.~Batra, ``{Habitat: A Platform
  for Embodied AI Research},'' in {\em 2019 IEEE/CVF International Conference
  on Computer Vision (ICCV)}, vol.~2019-Octob, pp.~9338--9346, IEEE, oct 2019.

\bibitem{Mnih2013}
V.~Mnih, A.~{Puigdom{\`{e}}nech Badia}, M.~Mirza, T.~Harley, T.~{P. Lillicrap},
  D.~Silver, and K.~Kavukcuoglu, ``{Asynchronous Methods for Deep Reinforcement
  Learning Volodymyr},'' {\em International Conference on Machine Learning},
  vol.~48, 2013.

\bibitem{Schulman2017}
J.~Schulman, F.~Wolski, P.~Dhariwal, A.~Radford, and O.~Klimov, ``{Proximal
  Policy Optimization Algorithms},'' {\em ArXiv}, 2017.

\bibitem{He2017}
K.~He, G.~Gkioxari, P.~Dollar, and R.~Girshick, ``{Mask R-CNN},'' in {\em 2017
  IEEE International Conference on Computer Vision (ICCV)}, vol.~2017-Octob,
  pp.~2980--2988, IEEE, oct 2017.

\bibitem{Jiang2018a}
J.~Jiang, L.~Zheng, F.~Luo, and Z.~Zhang, ``{RedNet: Residual Encoder-Decoder
  Network for indoor RGB-D Semantic Segmentation},'' {\em arXiv}, jun 2018.

\bibitem{Yang2019}
W.~Yang, X.~Wang, A.~Farhadi, A.~Gupta, and R.~Mottaghi, ``{Visual semantic
  navigation using scene priors},'' in {\em 7th International Conference on
  Learning Representations, ICLR 2019}, pp.~1--14, 2019.

\bibitem{Lyu2022}
Y.~Lyu, Y.~Shi, and X.~Zhang, ``{Improving Target-driven Visual Navigation with
  Attention on 3D Spatial Relationships},'' {\em Neural Processing Letters},
  vol.~54, no.~5, pp.~3979--3998, 2022.

\bibitem{Druon2020}
R.~Druon, Y.~Yoshiyasu, A.~Kanezaki, and A.~Watt, ``{Visual object search by
  learning spatial context},'' {\em IEEE Robotics and Automation Letters},
  vol.~5, no.~2, pp.~1279--1286, 2020.

\bibitem{Ye2021a}
J.~Ye, D.~Batra, A.~Das, and E.~Wijmans, ``{Auxiliary Tasks and Exploration
  Enable ObjectGoal Navigation},'' {\em Proceedings of the IEEE International
  Conference on Computer Vision}, pp.~16097--16106, 2021.

\bibitem{Chaplot2020b}
D.~S. Chaplot, D.~Gandhi, A.~Gupta, and R.~Salakhutdinov, ``{Object goal
  navigation using goal-oriented semantic exploration},'' {\em Advances in
  Neural Information Processing Systems}, vol.~2020-December, no.~NeurIPS,
  pp.~1--12, 2020.

\bibitem{Chaplot2020}
D.~S. Chaplot, D.~Gandhi, S.~Gupta, A.~Gupta, and R.~Salakhutdinov, ``{Learning
  to Explore using Active Neural SLAM},'' in {\em International Conference on
  Learning Representations (ICLR)}, apr 2020.

\bibitem{Ramakrishnan2022}
S.~K. Ramakrishnan, D.~S. Chaplot, Z.~Al-Halah, J.~Malik, and K.~Grauman,
  ``{PONI: Potential Functions for ObjectGoal Navigation with Interaction-free
  Learning},'' {\em Proceedings of the IEEE Computer Society Conference on
  Computer Vision and Pattern Recognition}, vol.~2022-June, pp.~18868--18878,
  2022.

\bibitem{Ramrakhya2022}
R.~Ramrakhya, E.~Undersander, D.~Batra, and A.~Das, ``{Habitat-Web: Learning
  Embodied Object-Search Strategies from Human Demonstrations at Scale},'' {\em
  Proceedings of the IEEE Computer Society Conference on Computer Vision and
  Pattern Recognition}, vol.~2022-June, pp.~5163--5173, apr 2022.

\bibitem{Radford2021}
A.~Radford, J.~W. Kim, C.~Hallacy, A.~Ramesh, G.~Goh, S.~Agarwal, G.~Sastry,
  A.~Askell, P.~Mishkin, J.~Clark, G.~Krueger, and I.~Sutskever, ``{Learning
  Transferable Visual Models From Natural Language Supervision},'' {\em
  Proceedings of the 38th International Conference on Machine Learning},
  vol.~139, pp.~8748--8763, feb 2021.

\bibitem{Yamauchi1997}
B.~Yamauchi, ``{Frontier-based approach for autonomous exploration},'' {\em
  Proceedings of IEEE International Symposium on Computational Intelligence in
  Robotics and Automation, CIRA}, pp.~146--151, 1997.

\bibitem{Wijmans2019}
E.~Wijmans, A.~Kadian, A.~Morcos, S.~Lee, I.~Essa, D.~Parikh, M.~Savva, and
  D.~Batra, ``{DD-PPO: Learning Near-Perfect PointGoal Navigators from 2.5
  Billion Frames},'' {\em arXiv}, nov 2019.

\bibitem{Huang2022a}
C.~Huang, O.~Mees, A.~Zeng, and W.~Burgard, ``{Visual Language Maps for Robot
  Navigation},'' {\em arXiv}, oct 2022.

\bibitem{Maksymets2021}
O.~Maksymets, V.~Cartillier, A.~Gokaslan, E.~Wijmans, W.~Galuba, S.~Lee, and
  D.~Batra, ``{THDA: Treasure Hunt Data Augmentation for Semantic
  Navigation},'' {\em Proceedings of the IEEE International Conference on
  Computer Vision}, pp.~15354--15363, 2021.

\bibitem{Chaplot2020a}
D.~S. Chaplot, R.~Salakhutdinov, A.~Gupta, and S.~Gupta, ``{Neural topological
  SLAM for visual navigation},'' in {\em Proceedings of the IEEE Computer
  Society Conference on Computer Vision and Pattern Recognition},
  pp.~12872--12881, 2020.

\bibitem{Li2022}
B.~Li, K.~Q. Weinberger, S.~Belongie, V.~Koltun, and R.~Ranftl,
  ``{Language-driven Semantic Segmentation},'' {\em arXiv}, pp.~1--13, jan
  2022.

\bibitem{Chen2022}
W.~Chen, S.~Hu, R.~Talak, and L.~Carlone, ``{Leveraging Large Language Models
  for Robot 3D Scene Understanding},'' {\em arXiv}, sep 2022.

\bibitem{Gadre2022}
S.~Y. Gadre, M.~Wortsman, G.~Ilharco, L.~Schmidt, and S.~Song, ``{CLIP on
  Wheels: Zero-Shot Object Navigation as Object Localization and
  Exploration},'' {\em arXiv}, pp.~1--22, mar 2022.

\bibitem{Khandelwal2022}
A.~Khandelwal, L.~Weihs, R.~Mottaghi, and A.~Kembhavi, ``{Simple but Effective:
  CLIP Embeddings for Embodied AI},'' in {\em 2022 IEEE/CVF Conference on
  Computer Vision and Pattern Recognition (CVPR)}, pp.~14809--14818, IEEE, jun
  2022.

\bibitem{Al-Halah2022}
Z.~Al-Halah, S.~K. Ramakrishnan, and K.~Grauman, ``{Zero Experience Required:
  Plug {\&} Play Modular Transfer Learning for Semantic Visual Navigation},''
  {\em Proceedings of the IEEE Computer Society Conference on Computer Vision
  and Pattern Recognition}, vol.~2022-June, pp.~17010--17020, 2022.

\bibitem{Min2022}
S.~Y. Min, Y.-H.~H. Tsai, W.~Ding, A.~Farhadi, R.~Salakhutdinov, Y.~Bisk, and
  J.~Zhang, ``{Object Goal Navigation with End-to-End Self-Supervision},'' {\em
  arXiv}, dec 2022.

\bibitem{Julia2012}
M.~Juli{\'{a}}, A.~Gil, and O.~Reinoso, ``{A comparison of path planning
  strategies for autonomous exploration and mapping of unknown environments},''
  {\em Autonomous Robots}, vol.~33, no.~4, pp.~427--444, 2012.

\bibitem{Sethian1996}
J.~A. Sethian, ``{A fast marching level set method for monotonically advancing
  fronts},'' {\em Proceedings of the National Academy of Sciences of the United
  States of America}, vol.~93, no.~4, pp.~1591--1595, 1996.

\bibitem{Liu2019}
Y.~Liu, M.~Ott, N.~Goyal, J.~Du, M.~Joshi, D.~Chen, O.~Levy, M.~Lewis,
  L.~Zettlemoyer, and V.~Stoyanov, ``{RoBERTa: A Robustly Optimized BERT
  Pretraining Approach},'' {\em arXiv}, jul 2019.

\bibitem{Anderson2018}
P.~Anderson, A.~Chang, D.~S. Chaplot, A.~Dosovitskiy, S.~Gupta, V.~Koltun,
  J.~Kosecka, J.~Malik, R.~Mottaghi, M.~Savva, and A.~R. Zamir, ``{On
  Evaluation of Embodied Navigation Agents},'' {\em arXiv}, jul 2018.

\end{thebibliography}

\end{document}